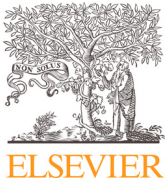
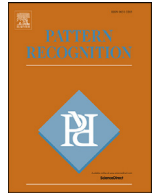

# Understanding and combating robust overfitting via input loss landscape analysis and regularization

Lin Li, Michael Spratling*

*Department of Informatics, King's College London, 30 Aldwych, London,WC2B 4BG, UK*

**ABSTRACT**

Adversarial training is widely used to improve the robustness of deep neural networks to adversarial attack. However, adversarial training is prone to overfitting, and the cause is far from clear. This work sheds light on the mechanisms underlying overfitting through analyzing the loss landscape w.r.t. the input. We find that robust overfitting results from standard training, specifically the minimization of the clean loss, and can be mitigated by regularization of the loss gradients. Moreover, we find that robust overfitting turns severer during adversarial training partially because the gradient regularization effect of adversarial training becomes weaker due to the increase in the loss landscape's curvature. To improve robust generalization, we propose a new regularizer to smooth the loss landscape by penalizing the weighted logits variation along the adversarial direction. Our method significantly mitigates robust overfitting and achieves the highest robustness and efficiency compared to similar previous methods. Code is available at https://github.com/TreeLLi/Combating-RO-AdvLC.

© 2022 The Author(s). Published by Elsevier Ltd.
This is an open access article under the CC BY-NC-ND license
(http://creativecommons.org/licenses/by-nc-nd/4.0/)

## 1. Introduction

Over the past decade, the progress of deep neural networks (DNNs) [1] has significantly improved the accuracy of machine intelligence in many pattern recognition tasks such as image recognition [2]. However, deep neural networks are vulnerable to adversarial attacks where artificial, human-imperceptible, perturbations are applied to the input space causing the accuracy of well-trained networks to easily be reduced to (almost) zero [3]. This issue, of adversarial vulnerability, has received considerable, and increasing, attention in the community [4]. Furthermore, a great concern has been raised in society regarding the safety of DNN-based systems as more and more such systems are deployed in the real world.

To date, adversarial training (AT) has been the most successful technique to improve the adversarial robustness of DNNs. However, adversarially robust generalization requires much more data [5] and robust training is therefore easier than standard training to overfit under the same data setting. Compared to benign overfitting [6] in standard training, adversarially robust overfitting is more problematic and can significantly harm robust performance [7,8]. Particularly, robust overfitting [7] is a form of overfitting where, even if the robust accuracy on the training data consistently increases, the test robustness drops after a certain epoch, usually around the first decay of the learning rate in the middle of training. Catastrophic overfitting [8] refers to the phenomenon in which robustness against multi-step adversaries suddenly collapses to zero over the course of a few epochs during training, while robustness against single-step adversaries soars to 100%. Robust overfitting is prevalent across multiple datasets and perturbation models, while catastrophic overfitting has, so far, only been demonstrated for single-step AT in normal adversarial settings.

This paper aims to explain the causes of overfitting in adversarial training, focusing on robust overfitting. This is achieved through analyzing how the loss landscape w.r.t. the input, specifically the first- and the second-order gradients (the slope and curvature, respectively), evolve during training. First, we observe that standard training, in common with adversarial training, suffers from the generalization issue of adversarial robustness and shares a similar pattern of loss gradients. We demonstrate that the robust generalization issue stems from minimizing the predictive loss on the clean input, which is implicitly implemented in adversarial training, and can be mitigated by either subtracting clean loss from adversarial loss or regularizing loss gradients. Moreover, we show that the effectiveness of training adversarial examples, and concomitantly the strength of the gradient regularization effect of adversarial training, decreases with the loss landscape's increasing curvature throughout training. The weakening regularization enables gradients to become larger, and hence, aggravates overfitting.

* Corresponding author.
 *E-mail addresses:* lin.3.li@kcl.ac.uk (L. Li), michael.spratling@kcl.ac.uk (M. Spratling).

To tackle robust overfitting, we propose a new regularizer, which is combined with adversarial training to smooth the loss landscape. The proposed method regularizes the weighted difference between the logit output of the clean input and its adversarial counterpart. The intuition here is to force an arbitrary point within the small radius around the input to have the same output as the input. Empirical result shows that this method effectively smooth the loss landscape and mitigates against robust overfitting. Compared to previous best practices, our proposal achieves the highest adversarial robustness of all assessed works and is significantly more computationally efficient.

## 2. Related work

Adversarial training is commonly formulated as a min-max optimization problem as shown in Eq. (1), where the inner loop searches for the strongest adversarial examples and the outer loop searches for the best parameters to minimize the loss on the generated adversarial examples.

$$\min_{\theta} \mathbb{E}_{(x,y) \sim D}[\max_{\delta \in \mathcal{B}(x,\epsilon)} \mathcal{L}(x+\delta, y; \theta)] \quad (1)$$

Fast gradient sign method (FGSM) and Projected gradient descent (PGD) are two representative methods for solving the inner maximization problem. FGSM [3] perturbs each pixel by a constant (budget) according to the sign of the gradients. PGD [9] performs a multi-step search, with each step using FGSM to perturb the result from the last step and then projecting the perturbation back to the constrained radius. From this perspective, FGSM can be considered to be PGD with only one step named PGD1. FGSM AT [3] is widely accepted as the least expensive training scheme due to its single-step nature. However, it suffers badly from catastrophic overfitting [8] a.k.a. label leaking [10] and gradient masking [11]. In contrast, multi-step PGD AT is much more effective [9] and seems immune from catastrophic overfitting [10,12]. Nevertheless, it is extremely time-consuming due to the iterative inner optimization process, and hence, scales poorly to large datasets. Recently, [7] report that PGD AT, albeit effective, also overfits in another manner called robust overfitting.

Initially, [8,13] attributed catastrophic overfitting to the rigid perturbation size of FGSM and [8,14,15] combined FGSM with random initialization to diversify the perturbation size and successfully prevent catastrophic overfitting. However, this result was later refuted by [16] who showed that it is the direction, not the size, of the perturbation that dominates catastrophic overfitting and as a result the methods proposed in [8,15], as well as PGD2 AT, suffer the overfitting problem given a larger perturbation budget or more training iterations. Besides, [13,16] observed that the loss surface w.r.t. the input becomes highly curved as a result of catastrophic overfitting. Consequently, single-step adversaries fail to accurately approximate optimal adversarial examples. In other words, the adversarial examples used during training become easier to classify.

Many approaches have been proposed to solve robust overfitting, but the cause is not fully explained. [7] tested several well-established approaches to prevent overfitting in standard training and found that none of them outperforms simply stopping early. However, subsequent empirical results show that robust overfitting can be mitigated by proper label and weight smoothing, a method called Knowledge Distillation with Standard-trained and Adversarially-trained self-teachers (KDSA) [17], and by data augmentation through a method called Consistency [18]. The correlation between the flatness of the weight loss landscape and the robust generalization gap is formally identified by [19] where they also propose an algorithm called adversarial weight perturbation to smooth the weight loss landscape. Recently, [20] found that robust overfitting is caused by learning to over-confidently predict some "hard" data points in the training set. The current work complements the existing literature by analyzing the input loss landscape which has never been done before in the context of investigating the cause of robust overfitting.

Our proposed regularizer belongs to the family of methods that smooth the loss landscape. Existing methods can be categorized as regularizing first-order (slope), second-order (curvature) and whole-order gradients. In practice, only first-order gradients can be directly constrained [21], while the others are too expensive to regularize directly. Therefore, modern approaches approximate higher-order gradients through proxies. The typical proxy is the distance (or dissimilarity) between the model's output in response to an input and its output to a neighboring example. By minimizing this distance the loss landscape is smoothed. The neighbor can be randomly sampled or searched for by (approximately) maximizing the aforementioned distance within a small radius around the input. The output here usually refers to the logits or the first-order gradients.

Particularly, CURE [22] penalizes the difference between the first-order gradients of the input and its FGSM-like adversarial example. GradAlign [16] minimizes the cosine dissimilarity between the first-order gradients of the input and its randomly sampled neighbor. LLR [23] searches for the neighbor where the linear approximation is maximally violated and penalizes the linear violation between the input and the neighbor as well as the the $l_1$ norm of first-order loss gradients. All these three methods require double backpropagation, which is computationally expensive. Hence, other methods try to avoid this expense by replacing the gradients with the outcome of a forward pass. LogitAlign [24] randomly samples the neighbor and penalizes the Kullback-Leibler (KL) divergence of the logits between the adversarial example generated for the input and the adversarial example generated for the neighbor. Consistency [18] generates the neighbor by augmentation and minimizes the Jensen-Shannon (JS) divergence of the output probabilities between the adversarial examples of the input and its neighbor. Both LogitAlign and Consistency require the generation of one additional adversarial example which can be extremely expensive if an iterative attack is used. RST [25] and UAT [26] search for the neighbor with the maximum KL divergence to the original example, and then minimizes the same KL divergence to enforce the smoothness. This iterative search pipeline makes it hard to scale especially when combined with adversarial training.

## 3. Revisiting the formulation of adversarial attack and training

This section reviews how gradients are exploited/sacrificed to affect the model's output, and suppressed/encouraged in adversarial training. In addition, it proposes the concept of adversarial effectiveness and develops a method to measure it. More importantly, it uncovers the correlation between the effectiveness of the training adversarial examples and the strength of implicit gradient regularization in adversarial training. We acknowledge that the theoretical analysis used here has been explored previously in other works [27–29] to derive conclusions such as the first- and second-order gradients are the two main sources of adversarial vulnerability, and to identify the correspondence between adversarial training and gradient regularization. Some of these previous insights are discussed in detail below, and are used to inspire or support our contribution. We extend the existing analysis to consider the interaction among gradients of different orders in adversarial attack and training and the connection between them.

The following notation is adopted: $x \in \mathbb{R}^d$ is a sample whose ground truth label is $y$ in dataset $D$ and is perturbed by $\delta \in \mathcal{B}(x, \epsilon)$ to fool the network. $\mathcal{B}(x, \epsilon)$ denotes the $\epsilon$-ball around the example $x$ with a specific distance measure ($\ell_\infty$ in this paper). $\delta_i$ is the perturbation applied along the dimension $x_i$. The network is parameterized by $\theta$ and the predictive loss is $\mathcal{L}(x, y; \theta)$ or $\mathcal{L}(x)$ for short. According to Taylor's theorem, adversarial loss can be approximated by the sum of clean loss and adversarial variance (AV):

$$\mathcal{L}(x+\delta) \approx \mathcal{L}(x) + \frac{1}{1!}\sum_i^d \nabla_{x_i}\mathcal{L}(x)(x_i + \delta_i - x_i)$$
$$+ \frac{1}{2!}\sum_{i,j}^d \nabla^2_{x_i x_j}\mathcal{L}(x)(x_i + \delta_i - x_i)(x_j + \delta_j - x_j)$$
$$\approx \mathcal{L}(x) + \underbrace{\sum_i^d \nabla_{x_i}\mathcal{L}(x)\delta_i + \frac{1}{2}\sum_{i,j}^d \nabla^2_{x_i x_j}\mathcal{L}(x)\delta_i\delta_j}_{\text{adversarial variance (AV)}} \quad (2)$$

where $\nabla_{x_i}\mathcal{L}(x)$ is the first-order gradient of $\mathcal{L}(x)$ w.r.t. the input variable $x_i$ and $\nabla^2_{x_i x_j}\mathcal{L}(x)$ refers to the second derivative of $\mathcal{L}(x)$ w.r.t. the input variables $x_i$ and $x_j$. From this perspective, robust accuracy increases with an increase in clean accuracy or a decrease in the magnitude of the loss gradients.

AV in the above equation is expanded only up to the second-order gradients for simplicity and computational feasibility, but the analysis below applies to the higher order gradients as well. It assumes that the network is twice differentiable. Because ReLU-networks are not twice differentiable, we validate our claims for ReLU networks experimentally, and the results are reported in A.2.

### 3.1. The effectiveness of adversarial attack

Since it has no impact on the clean loss $\mathcal{L}(x)$, an adversarial attack manipulates the perturbation to maximize adversarial loss (Eq. (2)) via the adversarial variance:

$$\max_{\delta \in \mathcal{B}(x,\epsilon)} \left[ \sum_i^d \nabla_{x_i}\mathcal{L}(x)\delta_i + \frac{1}{2}\sum_{i,j}^d \nabla^2_{x_i x_j}\mathcal{L}(x)\delta_i\delta_j \right] \quad (3)$$

Any non-zero gradient contributes to the adversarial vulnerability and, individually, a greater gradient magnitude implies a larger vulnerability. Attacks exploit (sacrifice) the vulnerability from a certain gradient by aligning (misaligning) it with its perturbation counterpart to make a positive (negative) contribution to adversarial variance. A gradient aligns with a perturbation when they have the same sign.

The theoretical upper bound of AV is:

$$\epsilon \sum_i^d |\nabla_{x_i}\mathcal{L}(x)| + \frac{\epsilon^2}{2}\sum_{i,j}^d |\nabla^2_{x_i x_j}\mathcal{L}(x)| \quad (4)$$

where each single perturbation $\delta_i$ reaches the budget $\epsilon$ and all gradients, while multiplied by their corresponding perturbation counterpart, contribute positively to adversarial variance. Note $|\cdot|$ denotes the absolute value. This only occurs when there is no conflict among gradients so that the perturbation is able to align with all gradients simultaneously:

$$\forall i \in \{1, ..., d\}: \text{sign}(\nabla_{x_i}\mathcal{L}(x)) = \text{sign}(\delta_i)$$
$$\forall i, j \in \{1, ..., d\}: \text{sign}(\nabla^2_{x_i x_j}\mathcal{L}(x)) = \text{sign}(\delta_i \delta_j) \quad (5)$$

As a simple example of gradient conflict consider the situation where the signs of the gradients $\nabla_{x_i}$, $\nabla_{x_j}$, and $\nabla^2_{x_i x_j}$ are positive, positive and negative respectively. No values for $\delta_i$ and $\delta_j$ can make $\nabla_{x_i}\mathcal{L}(x)\delta_i$, $\nabla_{x_j}\mathcal{L}(x)\delta_j$ and $\nabla^2_{x_i x_j}\mathcal{L}(x)\delta_i\delta_j$ all positive at the same time. Gradient conflict is common in practice, because $\delta_i$ is shared among gradients of differing-orders involving $x_i$.

When gradients conflict, some gradients must be sacrificed in order to exploit others. Optimal adversaries will be those that best trade-off between conflicting gradients, and the adversarial variance achieved by such optimal adversaries is the practical upper bound (Eq. (3)). The gap between the practical and theoretical bounds depends solely on the intensity of gradient conflict regardless of the specific adversary, and they are equivalent if there is no conflict.

Finding the optimal adversary is intractable. Hence, contemporary attack methods attempt to approximate it. A worse approximation produces a less effective adversarial example, for example by sacrificing a significant gradient to exploit a trivial one. Theoretically, the effectiveness of an adversarial example is measured by the current AV divided by its theoretical upper bound:

$$\mathbb{E}_{(x,y) \sim D}\left[ \frac{\mathcal{L}(x+\delta, y; \theta) - \mathcal{L}(x, y; \theta)}{\text{Eq. (4)}} \right] \quad (6)$$

Ideally, the term in the denominator should be the theoretical maximal adversarial variance, to reflect the intensity of the gradient conflict and the quality of approximation. However, solving both the theoretical (up to infinite order) and the practical upper bounds is infeasible, so in our experiments we adopt a reasonably strong attack, PGD50, to approximate the maximum (see A.1 for validation). This practical effectiveness measure (Eq. (7)) excludes the effect of gradient conflict and actually reflects the relative effectiveness relationship between the assessed attack and PGD50.

$$\mathbb{E}_{(x,y) \sim D}\left[ \frac{\mathcal{L}(x+\delta, y; \theta) - \mathcal{L}(x, y; \theta)}{\mathcal{L}(x+\delta^*, y; \theta) - \mathcal{L}(x, y; \theta)} \right] \quad (7)$$

where $\delta$ and $\delta^*$ are the perturbations generated by the assessed attack and PGD50 respectively. The effectiveness of the assessed attack becomes closer to that of PGD50 as this measure gets closer to 1.

### 3.2. Relating adversarial gradient regularization to attack effectiveness

Adversarial training optimizes the weights, $\theta$, to minimize adversarial loss, which can be translated into minimizing clean loss and adversarial variance (Eq. (2)). AV is the sum of many products of the gradient and the perturbation. To minimize one product, $\theta$ is updated towards increasing (decreasing) the gradient if its perturbation counterpart is negative (positive) and, consequently, the magnitude of the gradient increases if they align and decreases if not. Taking $\nabla_{x_i}\mathcal{L}(x)\delta_i$ as an example, its gradient with respect to the parameter is $\delta_i \nabla^2_{x_i \theta}\mathcal{L}(x)$. At each iteration, $\theta$ is updated by $-l\delta_i \nabla^2_{x_i \theta}\mathcal{L}(x)$ adding a variation $-l\delta_i [\nabla^2_{x_i \theta}\mathcal{L}(x)]^2$ in theory to the gradient $\nabla_{x_i}\mathcal{L}(x)$. The quadratic sub-gradient is always positive, and $l$ is the learning rate, so how the gradient varies depends only on the direction of the perturbation. From this perspective, adversarial training is equivalent to standard training plus a regularizer on all gradients:

$$\min_\theta \left[ \mathcal{L}(x) + \sum_i^d \nabla_{x_i}\mathcal{L}(x)\delta_i + \frac{1}{2}\sum_{i,j}^d \nabla^2_{x_i x_j}\mathcal{L}(x)\delta_i\delta_j \right] \quad (8)$$

The sign of the perturbation determines if the corresponding gradient is encouraged to reduce or increase, and the size of the perturbation determines the trade-off between gradient regularization and accuracy. We call this implicit gradient regularization effect "adversarial gradient regularization" to distinguish it from explicit gradient regularization [21,22,30].

The strength of adversarial gradient regularization, assuming gradients with larger magnitude have more space to regularize, is approximately measured by AV. Strength decreases, roughly speaking, with more gradients sacrificed or not fully exploited i.e. less effective adversarial examples.

## 4. The causes of robust overfitting

This section presents two accounts for robust overfitting through the analysis of input loss landscape. We first find that robust overfitting stems from the minimization of the clean loss and can be alleviated by gradient regularization. Secondly, we demonstrate that the training adversarial examples degenerate as the loss landscape curves, which weakens adversarial gradient regularization, and hence, aggravates robust overfitting. In summary, we argue that robust overfitting results from the clean optimization and relaxed adversarial gradient regularization.

We adopt input gradient norm (IG) and hessian spectrum (HS) to measure the slope and the curvature of the loss surface respectively. IG is the $\ell_1$ norm of first-order gradients, i.e. $||\nabla_x \mathcal{L}(x, y; \theta)||_1$, and HS is the $\ell_1$ norm of the top 20 eigenvalues of the Hessian. We used the power iteration method [28] to approximate the top-*N* eigenvalues by PyHessian [31], since directly computing the Hessian and its spectrum is computationally expensive and was beyond our available resources. The validity of using the top 20 eigenvalues to analyze curvature has been previously verified by [28,32]. We realize that the ReLU activation function in our models is not twice differentiable, which damages the faithfulness of HS as a measure of curvature. Nevertheless, additional experiments (A.2) verify that HS is still effective enough to indicate the variation of the curvature for ReLU-networks. We measure IG and HS for both the training and validation data, which are referred to using the subscripts $t$ and $v$ respectively in the following text. Note that while IG is calculated using the whole data set, due to computational resource constraints, HS is averaged over a subset of 2000 samples. The accuracy of this approximation is verified in A.3.

All experiments were performed using the following set-up unless specified otherwise. To allow comparison with related works, the models' architecture was wide-ResNet34. The dataset was CIFAR10 [33] with augmentations of random horizontal flip and random crop with 4 pixel padding. Stochastic gradient descent (SGD) was used to train models for 200 epochs with learning rate of [0.1, 0.01, 0.001], annealing at epochs 100 and 150. The momentum was 0.9, the weight decay was 1e-4, and the batch size was 100. The predictive loss was CrossEntropy. Experiments were run on Tesla V100 and A100 GPUs. The default perturbation budget $\epsilon$ for both adversarial training and evaluation was 8/255 under $\ell_\infty$ and the step size was $\epsilon/4$ for multi-step attacks. Adversarial robustness was evaluated by PGD50. FGSM-R AT refers to FGSM adversarial training with random initialization as [8] but the step size is still $\epsilon$.

### 4.1. Robust overfitting from clean optimization

We argue that standard training suffers from robust overfitting as well. We observe that adversarial vulnerability, measured by adversarial variance as discussed in Section 3.1, evolves similarly in standard and adversarial training. As shown in Fig. 1, for both standard and adversarial training, throughout the training process there is a consistent increase in the gap of adversarial variance between training and test data. This suggests that both training paradigms overfit worse in terms of adversarial robustness during training. Therefore, we argue that robust overfitting also exists in standard training. Note that the robust accuracy in standard training (which is close or equal to 0 for both data sets) does not appear to be like the conventional robust overfitting phenomenon because it is dominated by adversarial variance whose value is overwhelmingly greater than the corresponding clean loss value.

These results demonstrate that robust overfitting in adversarial training results from the minimization of clean loss i.e. standard training. Adversarial optimization is theoretically equivalent to the minimization of clean loss plus gradient regularization (Section 3.2), so standard training is implicitly implemented during adversarial training. In fact, robust overfitting appears milder in adversarial training compared to how it is in standard training according to Fig. 1. Moreover, these two training schemes present a similar pattern in the evolution of not only adversarial variance but also loss gradients: especially the novel divergent behavior of IG on the various sets after the first decay of the learning rate that raises a significant gap between adversarial vulnerability on training and test data.

To verify, we perform an ablation experiment by subtracting a portion of the standard training from FGSM-R AT ("FGSM-R AT - ST"). We train a set of models using the loss below with $\alpha$ ranging from 0 to 1.

$$\mathcal{L}(x + \delta) - \alpha \mathcal{L}(x) \quad (9)$$

As $\alpha$ was increased, the impact of clean optimization on the weight update was reduced and the gap between adversarial losses of training and test data i.e. the severity of robust overfitting was shrunk (see Fig. 2). This confirms our claim that there is a correlation between robust overfitting and clean optimization.

### 4.2. Regularizing gradients to mitigate overfitting

Robust overfitting can be mitigated by regularizing loss gradients. We hypothesize this based on the observation that adversarial training suffers from milder robust overfitting compared to standard training. To evaluate this hypothesis, we apply an input gradient regularizer, the squared $l_2$-norm of first order loss gradients, to standard training ("ST + IGR"):

$$\mathcal{L}(x) + \beta ||\nabla_x \mathcal{L}(x)||_2^2 \quad (10)$$

As can be seen in Fig. 2 as $\beta$ is increased, gradient regularization turns stronger and the gap between adversarial losses of training and test data is reduced more significantly. This hypothesis is further evaluated in the adversarial training setting in Section 5.2.

### 4.3. Relaxed gradient regularization in AT

Another account for robust overfitting in adversarial training is the relaxation of its implicit gradient regularization i.e. the regularization strength drops during training (as theoretically discussed in Section 3.2). This can be easily confirmed empirically: see Fig. 3, which shows a strong correlation between regularization strength and adversarial effectiveness. Specifically, it shows that the effectiveness of training adversarial examples declines consistently throughout training. Furthermore, adversarial examples, generated by the methods whose optimization ability is sensitive to the geometry of loss landscape, become less effective with the increase in the loss landscape's curvature. Common adversaries such as FGSM and PGD are all gradient-based, and thus, sensitive to some extent. Intuitively, growing curvature implies a wavier loss surface where the complexity of optimization is increased for gradient-based optimizers. Consequently, the quality of approximation drops if the optimizer is not effective at finding a solution in the wavy loss surface. Overall, the strength of adversarial gradient regularization decreases, so the underlying model overfits more severely regarding adversarial robustness, as the loss landscape becomes increasingly complicated.

Increased curvature has been observed before in catastrophic overfitting [16,22], but to the best of our knowledge, this is the first time it has been observed in adversarial training without catastrophic overfitting. This phenomenon is counter-intuitive to the claimed correspondence between adversarial training and gradient regularization [22]. The cause of this phenomenon is hardly explained in the existing literature. We argue that one possible account is, again, the relaxation of gradient regularization in adversarial training. As gradient regularization is relaxed, adversarial training becomes more similar to standard training so that adversarally-trained models become more like non-adversarially-trained models which have much larger HS (Fig. 1).

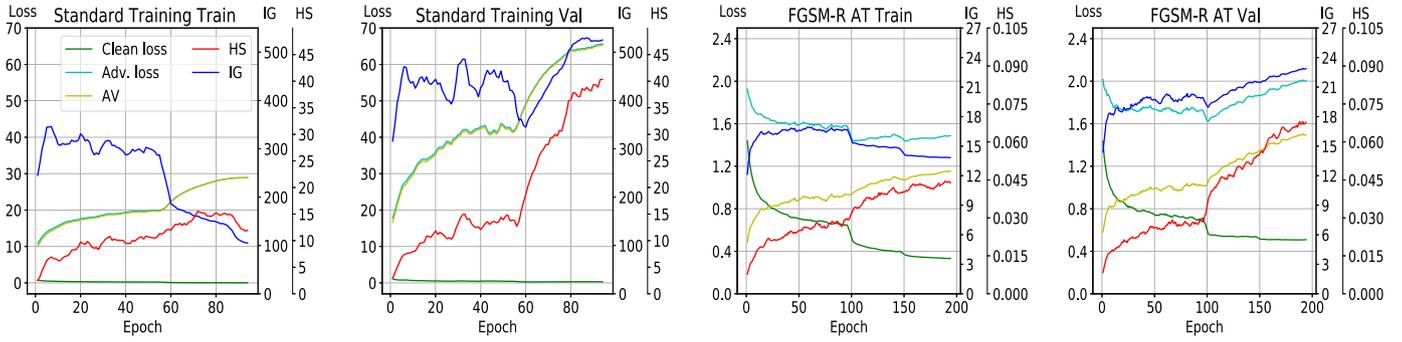

**Fig. 1.** IG, HS and loss for standard and adversarial training on validation and training data sets. Adversarial loss (Adv. loss) is computed based on the adversarial examples generated by PGD20, and adversarial variance (AV) is the difference between adversarial and clean loss (see Eq. (2)). Lines are smoothed with a moving average over 5 epochs for improved clarity.

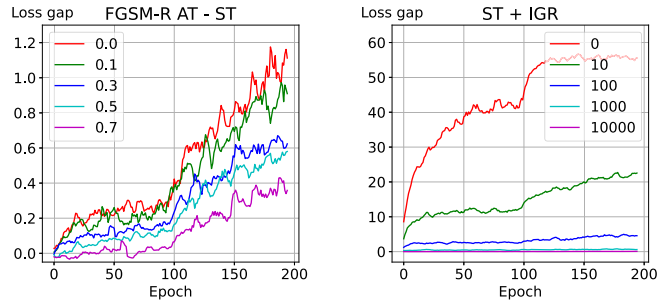

**Fig. 2.** How the gap between adversarial losses on training and test sets develops as training progresses. (left) FGSM-R AT with clean loss subtracted as in Eq. (9). (right) Standard training with input gradients regularized as in Eq. (10). Each line in the left (right) graph corresponds to a model trained with a different value of $\alpha$ ($\beta$) in Eq. (9) (Eq. (10)). Particularly, the training becomes FGSM-R AT and standard training when $\alpha$ and $\beta$ equals to 0 (red lines in both graph) which corresponds to the gap between the figures of adversarial losses in the Train and Val graphs of FGSM-R AT and standard training respectively in Fig. 1. Lines are smoothed over 5 epochs.

To verify the correlation between the effectiveness and the regularization strength, we simulate the degeneration of adversarial examples in training by manipulating the direction and the size of the perturbation. Specifically, we first generate a strong perturbation using PGD20 and then weaken it by either reducing the size of all perturbations by $p\%$, or by flipping the sign of $p\%$ (randomly sampled) perturbations. We then train an adversarially-trained robust model on those manipulated examples for one epoch to allow the gradients to change. As shown in Fig. 4a and Fig. 4b, HS negatively correlates to the effectiveness of the modified training examples in both cases.

Next, we test how the effectiveness of FGSM and PGD10 varies with the change of curvature. Two strategies are adopted to alter the curvature for a comprehensive validation. One is to reuse the models trained as described in the last paragraph (Fig. 4a and Fig. 4b). Apart from that, we randomly select a checkpoint during training and perturb its parameters by uniform noise $\mathcal{U}(-s, s)$ where $s \in \mathcal{U}(0, 1)$ (Fig. 4c). As shown in Fig. 4, the effectiveness of both adversaries declines approximately with the increase in the curvature in all cases. Moreover, FGSM's effectiveness drops more dramatically than PGD10's, indicating that the weaker adversary is more sensitive to the curvature.

## 5. Combating robust overfitting

The analysis in Section 4 provides the insight that robust overfitting in adversarial training is related to increasing curvature in input loss landscape. Therefore, we propose to alleviate robust overfitting by smoothing the input loss landscape. Directly regularizing curvature is, to our best knowledge, infeasible today, so our proposal regularizes the approximate curvature. Ideally, we want the original input example and any neighboring example, within a certain distance like $\epsilon$, to have the same output, i.e., a flat loss landscape:

$$\min_{\theta} \max_{\eta \in \mathcal{B}(x,\epsilon)} \rho(f(x+\eta;\theta), f(x;\theta)) \quad (11)$$

where $\rho(\cdot)$ can be any distance or dissimilarity measure. The above inner maximization searches for the perturbation $\eta$ so that the perturbed example differs most from the original example regarding the similarity between their output. This paradigm has been implemented before with various strategies as discussed in Section 2. Different from them, our method reuses the adversarial perturbation $\delta$ from the training pipeline to replace $\eta$ to save the expensive inner maximization:

$$\min_{\theta} \rho(f(x+\delta;\theta), f(x;\theta)) \quad (12)$$

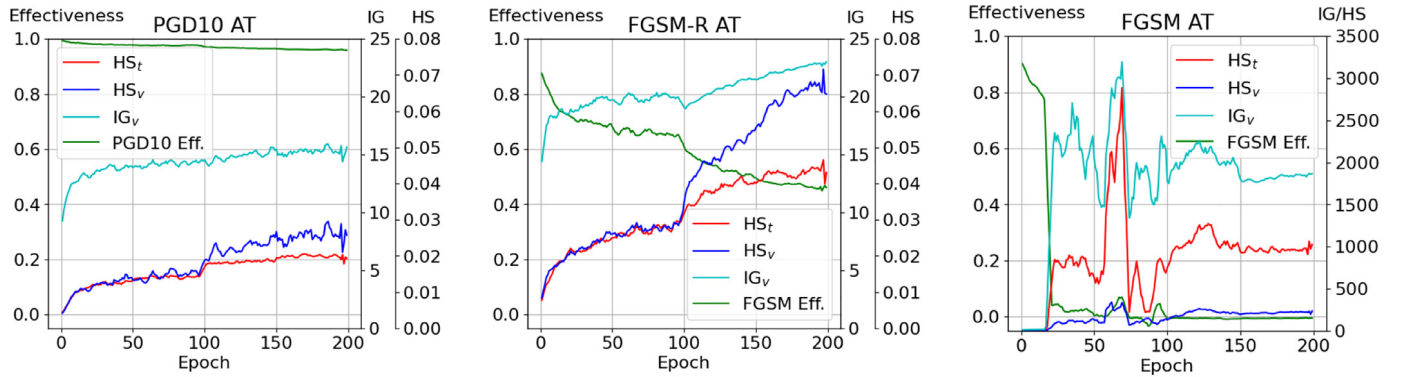

**Fig. 3.** HS, IG, and the effectiveness of adversarial examples (Eq. (7)) for various adversarial training schemes. The training adversaries' sensitivity to the loss surface's curvature increases from left to right and FGSM adversarial training (right figure) suffers catastrophic overfitting. PGD10 (FGSM) Eff. refers to the effectiveness of training adversarial examples generated by PGD10 (or FGSM). Lines are smoothed as in Fig. 1 and effectiveness is computed over 2000 samples like HS (A.3 for verification).

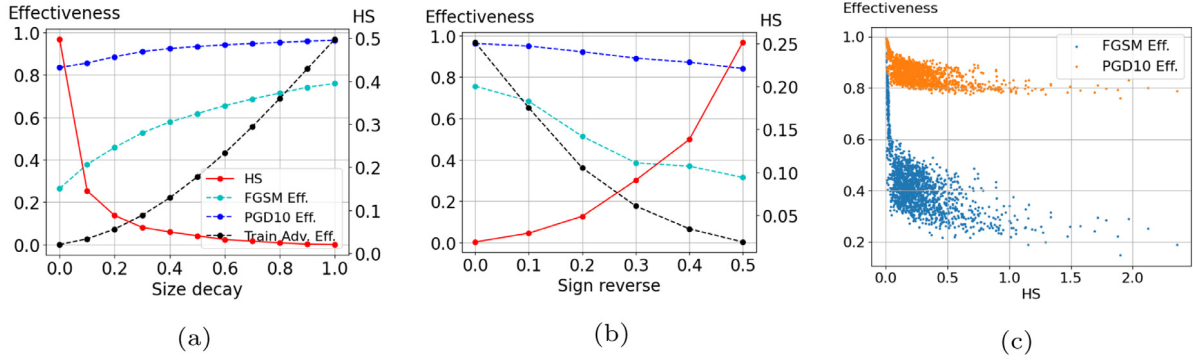

**Fig. 4.** HS and the effectiveness of FGSM and PGD10 for randomly modified models (a), and how curvature and adversaries' effectiveness vary as the size of perturbation decays (b), or the sign of perturbation flips (c). Each point in (a) represents a variant (model) of the selected adversarially-well-trained model. The X-axis in (b) is the multiplier applied to the perturbation, therefore, the perturbation becomes stronger and closer to the original perturbation from 0 to 1. The X-axis in (c) indicates the proportion of the perturbations whose sign is reversed, so 0 means no flip i.e. the original perturbation and 1 means all perturbations are flipped. The maximum reversal rate is limited to 0.5 since the reversed examples are no longer adversarial and catastrophic overfitting dominates the training once exceeded. Note that scales for HS in the subplots are different to clearly illustrate the trend. Train Adv. Eff. denotes the average effectiveness of the training examples during one-epoch of training.

Theoretically, $\delta$ is supposed to approximate $\eta$ well since the perturbation causing the largest loss variance ($\delta$'s objective) should also change the prediction logits significantly ($\eta$'s objective).

Specifically, our new smoothing method, which we call Adversarial Logits Consistency (AdvLC), penalizes the weighted sum of the difference between the logits for the clean input and its adversarial counterpart:

$$\mathcal{L}_{reg}(x_i) = \sum_{j}^{C} w_{i,j} o_{i,j} \quad (13)$$

$$o_{i,j} = |f_j(x_i + \delta; \theta) - f_j(x_i; \theta)| \quad (14)$$

$$w_{i,j} = \begin{cases} 1, & \text{if } o_{i,j} \text{ is top } N \text{ in the batch} \\ 0, & \text{otherwise} \end{cases} \quad (15)$$

where $o_{i,j}$ is the absolute difference for the $j$-th logit output for sample $x_i$, and is weighted by $w_{ij}$. Intuitively, the regularizer forces the model to output the same logits for both the clean and the corresponding adversarial example, i.e. no adversarial variance can be raised by attacks, so the loss landscape is flattened.

$w_{i,j}$ is introduced to allow control of the strength of the regularization for each single logit difference, $o_{i,j}$. In each batch, $w_{i,j}$ is made equal to 1 for each $o_{i,j}$ whose magnitude is in the top $N$ within the batch, and $w_{i,j} = 0$ otherwise. The aim is to alleviate the over-regularization on the tiny logit differences for a better trade-off between accuracy and robustness (see the comparison in Section 5.5 where the unweighted variant, $l_1$-norm, of our method shows a much worse trade-off between accuracy and robustness compared to the weighted version). Note that the top $N$ are selected among the entire mini-batch so it is possible for some examples to be totally unregularized. This is just one possible implementation of the proposed strategy. Other variants include setting all weights equal to one, which is equivalent to simply using the $l_1$-norm of the logit differences. Alternatively, setting $w_{i,j} = o_{i,j}$ would be equivalent to using the squared $l_2$-norm of the logit differences. Results for these alternatives are reported in Section 5.5.

We acknowledge that the proposed method is likely to be suboptimal and an improvement might be expected if $w$ was optimized for each difference term in an individual manner by some more advanced hyper-parameter optimization technique. However, the focus here is to demonstrate the effectiveness of smoothness regularization on mitigating robust overfitting, so we decide to leave this exploration for future work.

The overall objective is:

$$\mathcal{L}(x+\delta) + \lambda \mathcal{L}_{reg} \quad (16)$$

where $\lambda$ is the factor trading-off adversarial loss against regularization. Note that two forward passes, $f(x+\delta;\theta)$ and $f(x;\theta)$, are in practice shared between adversarial training and regularization, so the only extra computational overhead is in the calculation of the regularization loss. The implementation details are provided in Algorithm 1.

---

**Algorithm 1:** The proposed AdvLC regularizer showing how it is combined with arbitrary adversarial training. The size of a mini-batch is $M$.

**for** each batch **do**
    **for** $i = 1$ to $M$ **do**
        $\delta = \text{attack}(x_i, y_i)$
        $x'_i = \max(\min(x_i + \delta, 0), 1)$
        $o_{i,j} = |f_j(x_i) - f_j(x'_i)|$
        $\mathcal{L}_i = \mathcal{L}(x'_i, y_i)$
    **end**
    $w_{1,1}, ..., w_{M,C} = \text{weight}(o_{1,1}, ..., o_{M,C})$
    $\mathcal{L}_{reg} = \sum_{i,j}^{M,C} w_{i,j} o_{i,j}$
    $\mathcal{L}_{overall} = (\sum_i^M \mathcal{L}_i + \lambda \mathcal{L}_{reg})/M$
    $\theta = \theta - l \nabla_\theta \mathcal{L}_{overall}$
**end**

---

Our method by design improves on previous work (Section 2) in terms of both efficiency and effectiveness. First, compared to [16,21,22] the expensive computation of double backpropagation is reduced to one backward pass by replacing the first-order gradients with the logits. Second, compared to [16,18,22,23,25,26] adversarial examples are reused as the neighbor from the adversarial training pipeline to save extra forward and backward passes for the generation of neighbor examples. Third, regularizing logits variation along the adversarial direction, as well as the proposed weighting scheme, improves the trade-off between accuracy and robustness (Section 5.2).

*5.1. Benchmarking the proposed regularizer*

We follow the same setting as in Section 4 except for the the changes noted below. The model architecture was a PreActResNet18 [34]. The baseline single-step adversarial training method was FGSM-N AT [24] and the multi-step one was PGD10 AT. The robust accuracy was evaluated against the state-of-the-art attack AutoAttack [35] using Torchattacks [36] to exclude the false security of obfuscated gradients [12].

To train our method, $\lambda$ (Eq. (16)) was 0.5 (0.3) for FGSM-N (PGD10) AT combined with stochastic weight averaging (SWA) [17] and 0.55 (0.4) when not. $N$ (Eq. (15)) is fixed to 100 (i.e. top 10% given a batch size of 100 and 10 classes) for all training methods. The smoothness regularizers we compared to include IGR [21], CURE [22], GradAlign [16], LogitAlign [24] and Consistency [18] (see Section 2). Their strength was 1200 (1000), 1500 (1500), 7 (6), 100 (50) and 50 (50) in order for FGSM-N (PGD10) AT. For Consistency, the temperature $\tau$ was set to 1 (i.e. no temperature scaling) since we observed a false security regarding PGD robustness when using the original value of 0.5. Moreover, we fixed the distance between the input and neighbor in CURE to be $\epsilon \cdot \text{sign}(\nabla_x \mathcal{L}(x))$ since the original setting was less effective in our experiments. All the above hyper-parameters were optimized using grid search. Finally, we also compared against KDSA [17] (see Section 2) and its SWA combined version. This approach was implemented using the original configuration, except SWA was enabled from the $50^{th}$ epoch since this achieved higher robustness in our experiments.

We also validated the generalization ability of our method on TinyImageNet [37], which is a subset of large-scale natural image dataset ImageNet [2]. To stabilize the training, we linearly increased at each epoch the regularization strength, $\lambda$, of our method (or the $l_1$ norm variant of our method) from 0 at the start to 0.15 (or 0.07) at the end. The robust accuracy was evaluated against APGD50 [35] with 5 random starts, instead of AutoAttack because the latter attack required more GPU memory than we had available. Otherwise, the same settings were used as for the experiments with CIFAR10.

*5.2. Accuracy and robustness*

Our method, AdvLC, achieves the highest best and end robustness in both single- and multi-step AT groups in Table 1. The results are further improved when combined with SWA. Our method is the only one that improves single-step FGSM-N AT to outperform multi-step PGD10 AT. Furthermore our method enables single-step FGSM-N AT to produce higher robustness than multi-step PGD10 AT combined with the regularizors IGR, CURE, GradAlign and LogitAlign. Importantly, this dramatic robustness gain is not attained at the cost of a large reduction in accuracy compared to that obtained by other smoothing techniques. Compared to KDSA, our approach shows higher robustness both with and without SWA. However, KDSA exhibits greater accuracy compared to AdvLC and even the non-regularized baselines.

Almost all regularized training methods, except FGSM-N+IGR, produce lower IG and HS and a smaller gap between the best and end robust accuracy ("robustness diff." column) compared to the corresponding baselines, confirming their effectiveness in smoothing the loss landscape and mitigating robust overfitting. IGR is less effective in alleviating robust overfitting because $IG_t$, even non-regularized, drops at the later stage of training as observed in Fig. 1. GradAlign hardly reduces the gap between the best and end robustness and, even worse, makes the robustness less than that of the baseline. We attribute this ineffectiveness to the usage of Cosine Similarity since it only matches the gradient direction and ignores the variation in the gradient norm. All first-order gradient approaches, IGR, CURE and GradAlign, fail to improve the best robustness when combined with PGD10 AT suggesting that regularizing first-order gradients is insufficient to improve multi-step adversarial training. Last, we observe that all smoothness regularizers sacrifice accuracy for robustness.

**Table 2**
Performance of PGD10 AT and its AdvLC-regularized variants on TinyImageNet.

| Training | Accuracy | Robustness | Time |
|---|---|---|---|
| PGD10 AT | **44.71** | 19.74 | **1.143** |
| + AdvLC-L1 | 43.65 | 21.10 | 1.161 |
| + AdvLC-ours | 43.93 | **22.64** | 1.161 |

**Table 1**
Performance of various regularization methods combined with single and multiple step adversarial training on CIFAR10. Clean accuracy, and robust accuracy, were measured for the most robust, or "best", checkpoint and at the end of training. The best checkpoint was selected based on PGD5 robustness on the test set. The difference between those values gives an indication of the severity of overfitting. IG and HS were measured at the end of training on the test set. Time is an average measured for processing one mini-batch on a Tesla V100 in seconds. The best result is highlighted for each metric in each block. The results are averaged over multiple runs and the standard deviation is indicated by the value after the ± sign.

| Training | Accuracy (%) | | | Robustness (%) | | | IG | HS | Time |
|---|---|---|---|---|---|---|---|---|---|
| | best | end | diff. | best | end | diff. | | | |
| Standard | 44.25 ± 5.22 | 94.88 ± 0.18 | -50.63 | 2.61 ± 1.24 | 0.00 ± 0.00 | 2.61 | 276.0 | 17.22 | 0.017 |
| FGSM-N AT | 83.51 ± 0.15 | 83.72 ± 0.15 | **-0.21** | 41.77 ± 0.29 | 34.19 ± 0.13 | 7.58 | 48.00 | 0.230 | **0.047** |
| + IGR | 80.90 ± 0.48 | 83.56 ± 0.20 | -2.66 | 43.96 ± 0.20 | 35.98 ± 0.39 | 7.98 | 25.04 | 0.059 | 0.130 |
| + CURE | 81.32 ± 0.49 | 82.45 ± 0.46 | -1.14 | 44.18 ± 0.22 | 39.33 ± 0.06 | 4.84 | 18.87 | **0.018** | 0.162 |
| + GradAlign | 80.73± 0.32 | 81.35 ± 0.21 | -0.62 | 41.03 ± 0.52 | 33.95 ± 0.21 | 7.08 | 34.50 | 0.074 | 0.162 |
| + LogitAlign | 81.26 ± 0.28 | 82.04 ± 0.28 | -0.78 | 43.72 ± 0.39 | 38.38 ± 0.27 | 5.34 | 28.72 | 0.054 | 0.095 |
| + Consistency | 81.21 ± 0.37 | 82.75 ± 0.22 | -1.54 | 45.70 ± 0.27 | 43.53 ± 0.26 | 2.17 | 17.48 | 0.032 | 0.095 |
| + KDSA | **84.40** ± **0.15** | **85.20** ± **0.24** | -0.80 | 43.77 ± 0.31 | 42.15 ± 0.18 | 1.62 | 21.17 | 0.052 | 0.062 |
| + KDSA+SWA | 84.03 ± 0.33 | 85.17 ± 0.15 | -1.15 | 45.80 ± 0.08 | 45.04 ± 0.18 | **0.76** | 19.11 | 0.052 | 0.062 |
| + AdvLC | 81.06 ± 0.23 | 82.70 ± 0.24 | -1.64 | 47.07 ± 0.18 | 42.71 ± 0.41 | 4.36 | 20.14 | 0.049 | 0.054 |
| + AdvLC+SWA | 79.46 ± 0.19 | 82.14 ± 0.11 | -2.68 | **48.96** ± **0.09** | **48.14** ± **0.17** | 0.82 | **14.16** | 0.020 | 0.054 |
| PGD10 AT | 83.40 ± 0.21 | 83.58 ± 0.20 | **-0.19** | 46.50 ± 0.34 | 40.42 ± 0.22 | 6.08 | 30.90 | 0.117 | **0.326** |
| + IGR | 81.43 ± 0.28 | 83.50 ± 0.23 | -2.08 | 46.45 ± 0.14 | 41.90 ± 0.43 | 4.55 | 21.20 | 0.051 | 0.425 |
| + CURE | 80.65 ± 0.03 | 82.67 ± 0.28 | -2.02 | 46.24 ± 0.21 | 42.38 ± 0.81 | 3.86 | 17.90 | 0.020 | 0.445 |
| + GradAlign | 81.52 ± 0.54 | 82.28 ± 0.38 | -0.76 | 45.01 ± 0.45 | 38.96 ± 0.65 | 6.06 | 28.27 | 0.063 | 0.445 |
| + LogitAlign | 81.30 ± 0.27 | 81.93 ± 0.47 | -0.63 | 46.59 ± 0.03 | 43.42 ± 0.68 | 3.17 | 22.03 | 0.038 | 0.656 |
| + Consistency | 81.66 ± 0.27 | 81.95 ± 0.12 | -0.29 | 48.11 ± 0.22 | 47.66 ± 0.30 | 0.45 | 14.77 | 0.026 | 0.658 |
| + KDSA | **84.15** ± **0.18** | **85.15** ± **0.23** | -1.00 | 47.92 ± 0.07 | 46.15 ± 0.18 | 1.77 | 19.40 | 0.050 | 0.348 |
| + KDSA+SWA | 83.81 ± 0.55 | 84.73 ± 0.40 | -0.92 | 49.58 ± 0.14 | 49.26 ± 0.08 | **0.32** | 16.79 | 0.042 | 0.348 |
| + AdvLC | 81.16 ± 0.36 | 82.23 ± 0.40 | -1.07 | 48.73 ± 0.22 | 45.11 ± 0.43 | 3.62 | 18.03 | 0.033 | 0.337 |
| + AdvLC+SWA | 79.70 ± 0.20 | 82.09 ± 0.11 | -2.39 | **50.55** ± **0.33** | **49.73** ± **0.06** | 0.82 | **13.90** | **0.018** | 0.337 |

**Table 3**
Performance of FGSM-N AT and its AdvLC-regularized variants using different distance measures on CIFAR10. The format of this table is the same as, and described in the caption of, Table 1.

| Training | Accuracy (%) | | | Robustness (%) | | | IG | HS | Time |
|---|---|---|---|---|---|---|---|---|---|
| | best | end | diff. | best | end | diff. | | | |
| FGSM-N AT | **83.51** ± **0.15** | **83.72** ± **0.15** | **-0.21** | 41.77 ± 0.29 | 34.19 ± 0.13 | 7.58 | 48.00 | 0.230 | **0.047** |
| + $l_1$ | 81.09 ± 0.41 | 82.88 ± 0.16 | -1.80 | 46.01 ± 0.51 | 40.04 ± 0.90 | 5.97 | 26.00 | 0.146 | 0.054 |
| + KL | 80.83 ± 0.53 | 82.74 ± 0.05 | -1.91 | 46.29 ± 0.22 | 41.12 ± 0.40 | 5.17 | 25.58 | 0.075 | 0.055 |
| + JS | 81.18 ± 0.21 | 82.73 ± 0.32 | -1.55 | 46.69 ± 0.31 | 41.75 ± 0.39 | 4.94 | 23.76 | 0.061 | 0.055 |
| + Nuclear | 80.82 ± 0.19 | 82.64 ± 0.18 | -1.82 | 46.52 ± 0.25 | 42.51 ± 0.22 | **4.01** | 21.70 | 0.062 | 0.055 |
| + Squared $l_2$ | 80.73 ± 0.22 | 82.79 ± 0.51 | -2.06 | 47.00 ± 0.23 | 41.73 ± 2.10 | 5.28 | 21.72 | 0.058 | 0.054 |
| + ours | 81.06 ± 0.23 | 82.70 ± 0.24 | -1.64 | **47.07** ± **0.18** | **42.71** ± **0.41** | 4.36 | **20.14** | **0.049** | 0.054 |

*5.3. Efficiency*

Our method is the most efficient among all regularization approaches and adds only trivial computational overhead to the baselines as shown in Table 1. CURE and GradAlign are the most expensive methods in the single-step group, requiring 3 times as much compute time as ours due to the double backpropagation. When combined with multi-step AT, LogitAlign and Consistency are the most time-consuming ones, taking almost 2 times as long as ours, since they generate another adversarial example for regularization using the expensive training adversary PGD10. KDSA is also less efficient than the proposed method because it requires two more forward passes to get the output of two teacher models. Furthermore, KDSA in practice can be costlier than it appears in the table since two teacher models have to be well trained beforehand, and the time consumed in doing this is not counted in the reported times.

*5.4. Generalisation*

Our approach also generalizes well to the alternative dataset TinyImageNet. As shown in Table 2, it significantly improves the robustness of the baseline PGD10 AT with only a trivial additional computational cost. Besides, the proposed weight scheme consistently boosts the trade-off between accuracy and robustness when compared to the non-weighted version of our method i.e. $l_1$ norm variant (see Section 5).

*5.5. Ablation analysis*

The effects of the specific form of regularisation method used, Eq. (13) to Eq. (15), was explored and the results are shown in Table 3. It can be seen that the proposed weight scheme achieves the best trade-off between accuracy and robustness among all the evaluated distance measures while preserving the same efficiency. Our method attains a considerably higher robustness, and a similar accuracy, compared to the $l_1$-norm. It achieves a slightly better accuracy than the squared $l_2$-norm, and a similar robustness. It outperforms KL and Nuclear regarding both accuracy and robustness. It improves the robustness of JS by a relatively considerable amount while its accuracy is just slightly behind JS's. It was observed that the results for both $l_1$-norm and squared $l_2$-norm were more varied than those for the proposed method. This can be seen in the higher standard deviation in the results particularly at the end of training. They can be stabilized, at the cost of poorer robustness, by weakening the regularization. Overall, the performance gain of the proposed method, although marginal, validates our choice of regularization method (see Section 5). Moreover, by comparing the results in Table 3 with those in Table 1 it can be seen that regardless of what distance measure is used, the proposed regularization method consistently produces higher robustness with single-step adversarial training than all previously proposed methods.

## 6. Conclusion

This work contributes towards understanding the overfitting mechanism and improving the robust generalization of adversarial training. First, we theoretically discuss the function of input loss gradients in adversarial attack and training, and the correlation between the effectiveness of training adversarial examples and the strength of implicit gradient regularization in adversarial training. We then analyze how the slope and the curvature of input loss landscape evolve during adversarial training particularly when robust overfitting occurs. We find that robust overfitting (1) stems from the minimization of clean loss; (2) turns severer with the decrease in the strength of gradient regularization as a result of increasing curvature in the input loss landscape and the degeneration of training adversarial examples. Based on these insights, we hypothesize and verify that robust overfitting can be mitigated by smoothing the loss landscape. Last, we propose a new smoothing method, AdvLC, that regularizes the logits variance along the adversarial direction in a weighted way. It outperforms existing methods significantly in terms of both robustness and efficiency.

One limitation of this work is that we have not explained why optimizing clean loss leads to robust overfitting especially the novel divergent behavior of IG on seen and unseen data. Another limitation is that the proposed method, like all other smoothness regularizers, sacrifices accuracy for robustness. Future work might usefully explore these two problems to enhance understanding about robust overfitting and to further improve the performance of adversarial training.

**Declaration of Competing Interest**

The authors declare that they have no known competing financial interests or personal relationships that could have appeared to influence the work reported in this paper.

**Data availability**

Data will be made available on request.

**Acknowledgement**

The authors acknowledge the use of the research computing facility at King's College London, King's Computational Research, Engineering and Technology Environment (CREATE), and the Joint Academic Data science Endeavour (JADE) facility. This research was funded by the King's - China Scholarship Council (K-CSC).

**Appendix A. Empirical verification of approximations**

This section empirically verifies several approximations used in the main text.

*A1. PGD convergence*

We conduct experiments to show that PGD50, used in computing the effectiveness of adversarial examples, is a good approximation of the optimal adversary. Empirical result suggests that 50 steps is enough for the optimizer, projected gradient descent (PGD), to converge within the $\epsilon$-ball. As shown in Fig. A1, both robust accuracy and adversarial loss reach steady-state values once the number of steps is increased to 50 or more. The exception is the loss on the validation set which continues to grow until the maximum number of iterations tested i.e. 200. However, this exception does not affect our analysis since in this work all adversarial examples' effectiveness is computed with respect to the training data. We acknowledge the existence of other more advanced adversaries like [35] outperforming PGD, in other words, potentially offering a more faithful approximation. Nevertheless, we choose PGD50 since the gap between the adversarial losses achieved is trivial and PGD is easier to implement and widely accepted in many other works.

*A2. HS validity*

We first illustrate how the non-twice-differentiability of ReLU-networks affects the value of Hessian. The intermediate calculation between the loss function and the input consists of:

$$\mathcal{L}(x, y; \theta) = \mathcal{L}(s(g(x)), y; \theta) \tag{A.1}$$

where $s(\cdot)$ denotes the softmax layer and $g(\cdot)$ refers to the whole network whose outputs are logits. The first-order gradients of the loss w.r.t. the input, using chain rule, are:

$$\frac{\partial \mathcal{L}}{\partial x} = \frac{\partial \mathcal{L}}{\partial s} \frac{\partial s}{\partial g} \frac{\partial g}{\partial x} \tag{A.2}$$

and the second-order gradients, using the product rule, are:

$$\frac{\partial^2 \mathcal{L}}{\partial x^2} = \frac{\partial [\frac{\partial \mathcal{L}}{\partial s} \frac{\partial s}{\partial g}]}{\partial x} \frac{\partial g}{\partial x} + \frac{\partial \mathcal{L}}{\partial s} \frac{\partial s}{\partial g} \frac{\partial \frac{\partial g}{\partial x}}{\partial x} \tag{A.3}$$

Assuming modern ReLU-networks, $g(\cdot)$ is a piecewise linear function made up of ReLU activation and linear layer, so its second-order gradients are zeros and the above equation reduces to:

$$\frac{\partial^2 \mathcal{L}}{\partial x^2} = \frac{\partial [\frac{\partial \mathcal{L}}{\partial s} \frac{\partial s}{\partial g}]}{\partial x} \frac{\partial g}{\partial x} \tag{A.4}$$

However, we know that, although the second derivatives are zero, the first-order gradients of $g(\cdot)$ change dramatically at the points where the input to ReLU equals zero. This reduced Hessian (Eq. (A.4)) thus does not reflect the full curvature of the loss surface.

To verify the validity of HS in measuring the loss surface's curvature for ReLU-networks, we compare the HS from ReLU-networks ($HS_{relu}$) with the gradient alignment from the same networks ($GA_{relu}$), and the HS from dual-softplus-networks ($HS_{sp}$). GA measures the curvature indirectly through the similarity between the first-order gradients of the input and its randomly perturbed neighbor within the $\epsilon$-ball:

$$\mathbb{E}_{(x,y)\sim D, \eta \sim \mathcal{U}([-\epsilon,\epsilon]^d)}[\cos(\nabla_x \mathcal{L}(x, y; \theta), \nabla_x \mathcal{L}(x+\eta, y; \theta))] \tag{A.5}$$

where $\cos(\cdot, \cdot)$ returns the cosine similarity between two inputs. Calculating GA has the advantage that it does not involve second-order gradients. A higher GA indicates a more linear loss surface i.e. a smaller curvature. Another control is to compute HS for the dual network with all ReLU units replaced by softplus units. Softplus transforms the input as:

$$\text{softplus}(x) = \log(1 + e^x) \tag{A.6}$$

and can be considered as a twice differentiable approximation to ReLU. Therefore, HS should fully reflect the curvature for softplus networks. All the remaining hyper-parameters were kept the same for a fairer comparison. The empirical results confirm that HS is a valid indicator of the loss surface's curvature, because the trend of $HS_{relu}$ aligns well with both $GA_{relu}$ and $HS_{sp}$ throughout adversarial training, as shown in Fig. A2.

*A3. Sampling faithfulness*

Metrics like HS and the effectiveness of FGSM and PGD10 were averaged over 2000 samples instead of the whole dataset due to constraints on computational resources and time. Fig. A3 shows how the values of these approximations vary as more samples are used in the computation. The effectiveness values for both adversaries are very stable and the results are almost identical when measured using 2k samples and 10k samples (i.e. the whole validation set). In contrast, the estimated value of HS fluctuates acutely when the number of samples is small. Nevertheless, averaging over 2000 samples ensures the deviation from the true value (averaged over the whole dataset) to be reasonably small ($\pm 10\%$). Therefore, the number 2000 was chosen based on the balance between efficiency and the faithfulness of the approximation.

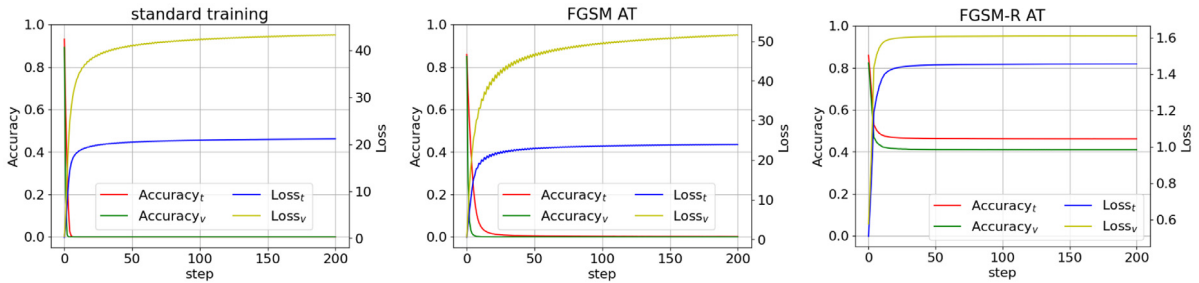

**Fig. A1.** How robust accuracy and adversarial loss vary with the increase in the number of steps in PGD for various datasets and training schemes.

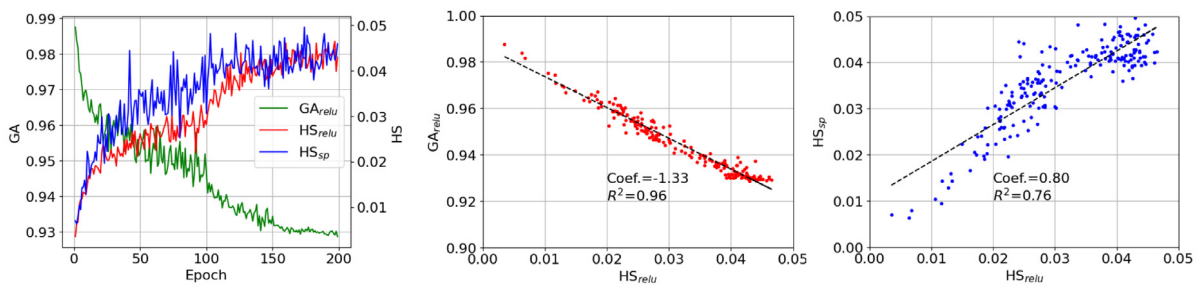

**Fig. A2.** Correspondence, measured using the training data, between the HS from ReLU-networks and (a) the gradient alignment for the same networks, and (b) HS from softplus-networks.

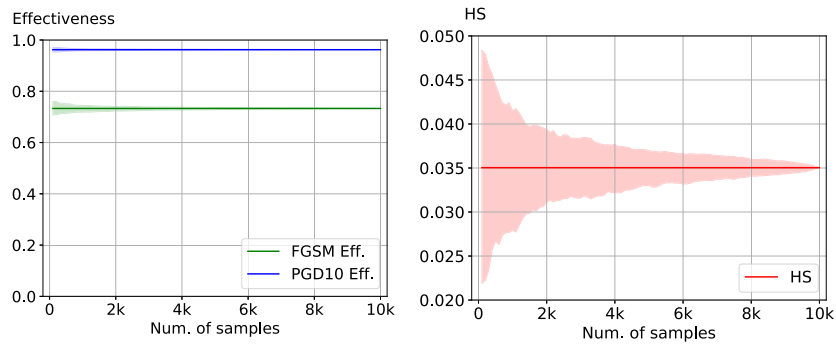

**Fig. A3.** Sensitivity of measurements as a function of the number of samples being averaged. The solid line indicates the true value of the measurements which is averaged over 10k samples (i.e. the whole validation set). The marked area is determined by the minimum and maximum of the estimation over the given number of samples in 10k trials.

**Lin Li** received a M.Sc. degree in computing from Imperial College London. He is currently a Ph.D. student in computer science at the Department of Informatics, King's College London. His research interest includes adversarial robustness and interpretability of deep learning

**Michael Spratling** received a B.Eng. degree in engineering science from Loughborough University and M.Sc. and Ph.D. degrees in artificial intelligence and neural computation from the University of Edinburgh. He is currently Reader in Computational Neuroscience and Visual Cognition at the Department of Informatics, King's College London. His research is concerned with understanding the computational and neural mechanisms underlying visual perception